\documentclass[conference]{IEEEtran}
\usepackage[pdftex]{graphicx}
\usepackage[numbers]{natbib}
\usepackage{fixltx2e}

\ifCLASSINFOpdf
\else
\fi
\begin{document}

%
\title{Evaluation of Joint Multi-Instance Multi-Label Learning For Breast Cancer Diagnosis}
\author{\IEEEauthorblockN{Baris Gecer\IEEEauthorrefmark{1},
Ozge Yalcinkaya\IEEEauthorrefmark{2}, Onur Tasar\IEEEauthorrefmark{3} and
Selim Aksoy\IEEEauthorrefmark{4}}
\IEEEauthorblockA{Department of Computer Engineering,\\
Bilkent University,
Ankara, 06800, Turkey\\
\IEEEauthorrefmark{1}baris.gecer@cs.bilkent.edu.tr,
\IEEEauthorrefmark{2}ozge.yalcinkaya@cs.bilkent.edu.tr,
\IEEEauthorrefmark{3}onur.tasar@cs.bilkent.edu.tr,
\IEEEauthorrefmark{4}saksoy@cs.bilkent.edu.tr}}

\maketitle

\IEEEpeerreviewmaketitle

\begin{abstract}
Multi-instance multi-label (MIML) learning is a challenging problem in many aspects. Such learning approaches might be useful for many medical diagnosis applications including breast cancer detection and classification. In this study subset of digiPATH dataset (whole slide digital breast cancer histopathology images) are used for training and evaluation of six state-of-the-art MIML methods.

At the end, performance comparison of these approaches are given by means of effective evaluation metrics. It is shown that MIML-kNN achieve the best performance that is $\%65.3$ average precision, where most of other methods attain acceptable results as well.

\end{abstract}

\section{Introduction}
Breast cancer is the most prevalent form of cancers among women~\cite{veta2014breast}. Although those who live in the developed world have more survival rates~\cite{boyle2008world}, patients have less change in developing countries. Medical image processing might have a huge contribution to experts for analysis of histopathology images by improving interpretation or indicating candidate disease locations~\cite{veta2014breast}. 

\begin{figure}[h]
\center
\label{fig:cancer}
\includegraphics[width=3.4in]{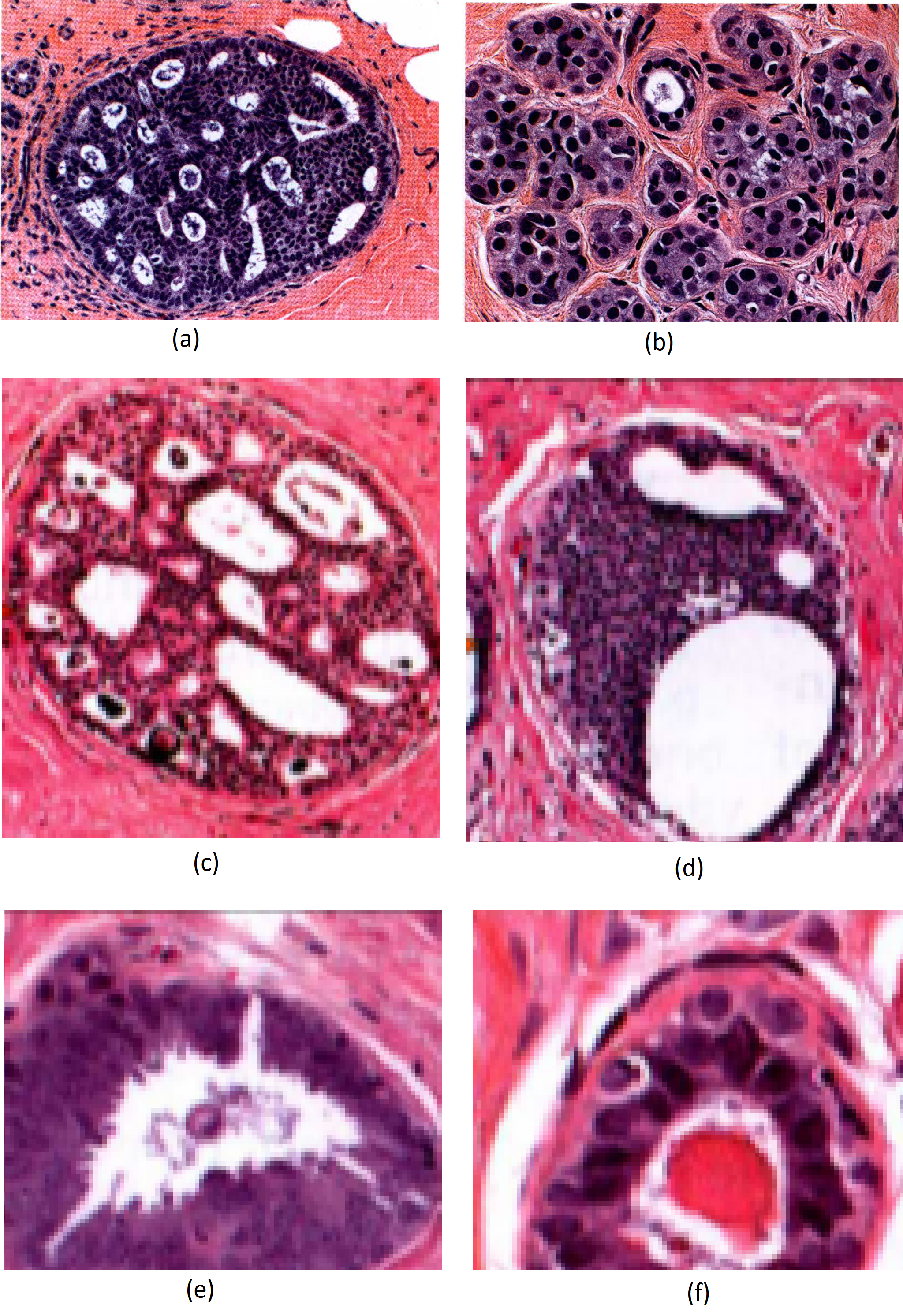}

\center
\caption{Some of the breast cancer types: (a) Atypical ductal hyperplasia. (b) Atypical lobular hyperplasia. (c) Ductal carcinoma in situ. (d) Atypical hyperplasia. (e) proliferative disease. (f) Non proliferative Changes Only}
\end{figure}

Some different types of breast cancers on histopatholgy images are shown in Fig.~\ref{fig:cancer}. Breast cancers does not have a particular shapes and their types are not necessarily consecutive. This make its detection and classification very hard for both experts and computer aid systems. There are so many information about breast cancer and its types in the literature but we omit the informations that require expertise in this field and focus on learning from labelled data by the experts in this study.

We introduce a well-collected dataset of digital histopathology images: digiPATH project in which consists of 240 digital images (in average size of 90.000$\times$70.000 pixels) of breast biopsies from 14 different diagnostic categories of cancer~\cite{mercan2014localization}. Each H\&E stained biopsy slides which were scanned at 40X magnification, labeled by 203 pathologists where 3 of them are world-class expert in this area. Rectangular part of the image that is visible on the pathologist's screen are logged while they diagnosis the images where world-class experts marked a bounding box in each image to indicate disease.

Diagnosis of a particular histopathology image might differentiate among experts and there might be more than one label from an expert for an image as it can be seen in the dataset. Thus, detection and classification of a case become more complicated than the traditional machine learning aspect, as we know one digital image might have healthy and different type of diseased regions at the same time. Such data set can be used for machine learning by only combining multi-instance and multi-label learning approaches.

\begin{figure}[t]
\center
\label{fig:figure}
\includegraphics[width=3.7in]{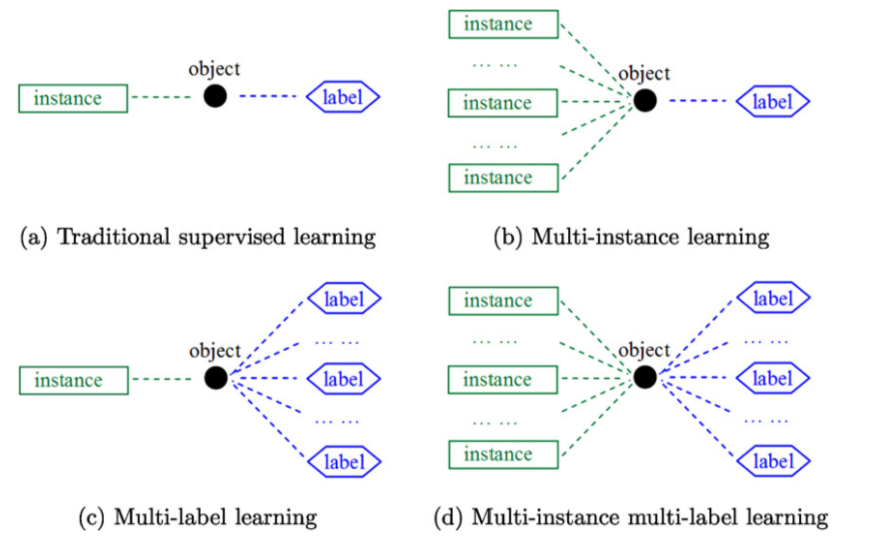}

\center
\caption{\normalsize{Illustration of traditional, multi-instance, multi-label and multi-instance multi-label learnings in a nutshell. This figure is obtained from~\cite{zhou2006multi}}}
\end{figure}

	Multi instance learning is a variation of the supervised learning. In classical supervised learning (Fig.\ref{fig:figure}a),  each sample in the training set is assigned to a specific label. 
	
	In multi-instance learning as illustrated in Fig.\ref{fig:figure}b, each image has one or more regions of interest associated with a label. So, each training image is considered as a bag of instances. In binary case, the image is labeled as positive if there is at least one positive instance in the bag. It is labeled as negative if all the instances are negative~\cite{yang2005review}. Goal of the multi-instance learning is to classify test images using the trained classifier by multi-instance images.

	Multi-label learning is another variation of the supervised learning where each image has multiple labels instead of only one as shown in Fig.\ref{fig:figure}c. For example, a nature image can be labeled as any of objects that is included in the image (i.e. sea, island and sky). The goal is to train a classifier from a collection of these multi-labeled images.

	One can combine these two learning methods for learning from such images that has a bag of instances and each image is associated with one or more labels (Fig.\ref{fig:figure}d).	In our research, the main interest is such combination which is the most convenient to solve the problem mentioned above where each histopathology image has multiple concerned regions and labeled by the pathologists individually. 
	
	This paper is organized as follows: in Section 2 we provide an overview of related work. Section 3 provides a detailed explanation of the experiments done and effectiveness of the methods are demonstrated. Finally, we provide a discussion in Section 4 and draw conclusions in Section 5.
	
\section{Related Work}
\label{sec:rw}
During the past few years, many MIML algorithms have been introduced. To name a few of state-of-the-art MIML methods, Zhou and Zhang~\cite{zhou2006multi} proposed MIMLSVM by decomposing MIML to single-instance multi-label learning and MIMLBOOST by decomposing MIML to multi-instance single-label learning. Zhang et.al.~\cite{zhang2008m3miml} proposed M3MIML, (i.e. Maximum Margin Method) which learns from multi-instance multi-label examples by maximum margin strategy and an adaptation of radial basis function (RBF) method~\cite{bishop1995neural} MIMLRBF which is a neural network style algorithm is proposed by Zhang et.al.~\cite{zhang2009mimlrbf}. A MIML nearest neighbor method MIML-kNN is proposed by Zhang and Min-Ling~\cite{zhang2010k}, that uses the popular k-nearest neighbor techniques. Yu-Feng Li et.al.~\cite{li2012towards} proposed KISAR (Key Instances Sharing Among Related labels) which is a MIML algorithm tries to observe the relation between instance and label by considering the fact that highly relevant labels share some instances.

\section{Experiments}
\begin{figure*}[t]
\center
\includegraphics[width=7.2in]{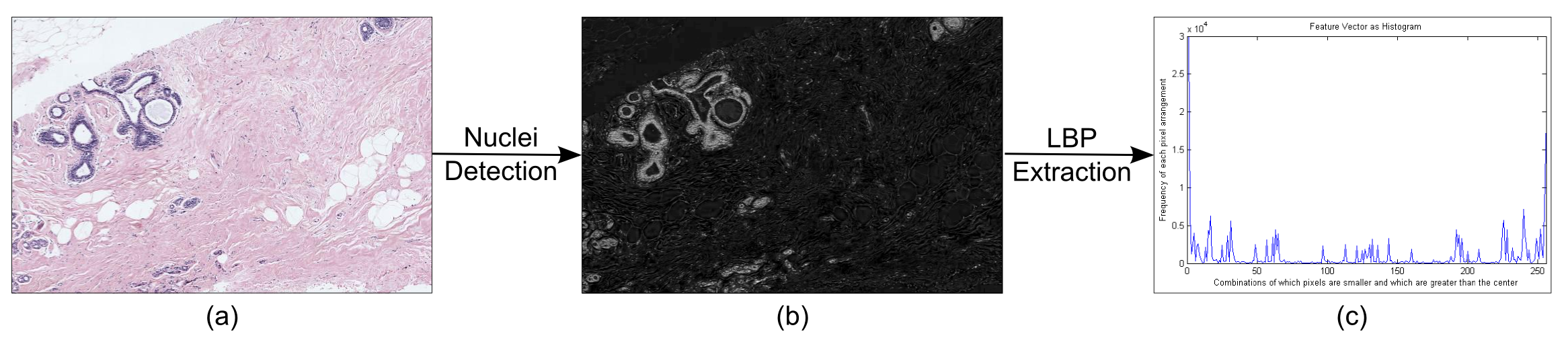}

\center
\caption{(a) Region of interest that is densely invested by an expert. This region is cropped from the whole slide image according to the view-log data by the method proposed by~\cite{mercan2014localization}. (b) Extracted nuclei of (a) in gray-scale with a transformation based on~\cite{ruifrok2001quantification}. (c) Histogram of LBP features of (b) with 8 neighbour setup.}
\label{fig:feature}
\end{figure*}

In this study, we compare the effectiveness of state-of-the-art algorithms that combine multi-instance and multi-label learning on a subset of digiPATH project dataset which is mentioned earlier. 

Firstly, we extract LBP features from regions that consist dense information about existence and type of a cancer from the behaviour of experts. The methods mentioned in section.~\ref{sec:rw}, then, trained and tested with these features. Finally performance of these methods are evaluated.

\subsection{Feature Extraction}

For brevity, we use only 120 images from the set and their corresponding diagnoses that are labelled by 3 world-class experts. Note that each expert might label a particular case with more than one diagnosis. View-logs of these experts are used to detect region of interests (ROI) of the images by the proposed method in the study \cite{mercan2014localization}, where these ROIs are detected by considering zoom peaks, slow pannings and fixation durations that are longer than 2 seconds. 

Most of the time mutual spatial arrangement of nuclei in a tissue is enough information for detection of a breast cancer and its classification to a cancer type. Thus, we first apply a transformation based on the method in~\cite{ruifrok2001quantification} to cropped ROI images (i.e. Fig.~\ref{fig:feature}(a)) that are originally in RGB color space where the transformation generate a map (i.e. Fig.~\ref{fig:feature}(b)) that unveil nuclei. Then we extract LBP features (i.e. Fig.~\ref{fig:feature}(c)) of this gray-scale nuclei image by comparing each pixel to each of its 8 neighbours and normalize features such that they add up to 1. Only 5 classes of disease types over 14 classes, are considered as labels in the experiments that are 'Non proliferative changes only', 'Usual ductal hyperplasia', 'Atypical ductal hyperplasia', 'Ductal carcinoma in situ', 'Invasive carcinoma'.

In the experiments we consider a bag of features of detected ROIs and labels of a particular expert on a particular image as a 'case'. We divide 360 cases (120 images and 3 experts) into well-balanced training and testing sets with respect to their disease types. When we remove cases that does not include any significant ROI, training and testing sets end up with 173 and 153 cases respectively.
\subsection{Methods}
We experimented six algorithms on our dataset and evaluated results according to some popular evaluation criteria. All of the algorithms return a predicted label set after testing phase. We used this predicted label set and the actual test label set while doing evaluations. Short description of algorithms are as follows:

MIML-kNN: This algorithm solves MIML problem by using the popular k-nearest neighbor techniques. Given a set of MIML training examples, it considers their neighbors and citers which regard it as their own neighbors. The MIML-kNN algorithm involves two different parameters, i.e. r (the number of nearest neighbors considered) and c (the number of citers considered). According to experiments in~\cite{zhang2010k}, we selected those parameters as r = 10, c = 20.

MIMLRBF: In this algorithm, two layer neural network structure is used. This neural network gets a bag X consisting of n instances {x\textsubscript{1}, x\textsubscript{2}, . . . , x\textsubscript{n}}, where each instance x\textsubscript{k} is a d-dimensional feature vector and each output unit of it is related to a possible class label. The first layer of this neural network is constituted by using k-medoids clustering. After that, the weights between first and second layer are computed. Finally, the test set is given to the trained neural network for prediction. MIMLRBF algorithm involves two different parameters, i.e. the fraction parameter α and the scaling factor $\mu$. We chose those parameters as $\alpha$ = 0.1, $\mu$ = 0.6 according to experiments in~\cite{zhang2009mimlrbf}.

MIMLSVM: This algorithm decomposes the multi instance multi label problem into single instance single label problems instead of considering the problem as a whole. Let X denotes the bag of instances and Y the set of class labels. MIMLSVM maps the bag of instances X\textsubscript{i} into a single instance z\textsubscript{i} with the help of constructive clustering. In this step, the algorithm reduces the problem to single instance multi label problem. After that MIMLSVM produces a number of independent single instance single label components by using MLSVM algorithm~\cite{zhou2006multi}. Parameters of the MIMLSVM are type of svm, gamma value of svm, cost and ratio; we chose type of the svm as RBF and gamma as 1; ratio is the number of clusters, cost parameter is the value used for the base svm classifier, these values were selected as 0.2 and 1, respectively.

MIMLBOOST: This is another algorithm that reduces the problem into simpler ones like MIMLSVM. The algorithm decomposes the multi instance multi label problem into multi instance single label problems by converting each MIML examples to MISL one. After this step intermediate MIBOOSTING algorithm reduces the MISL problems to single instance single problems~\cite{zhou2006multi}. The algorithm takes three parameters; round, svm, cost. Svm and cost parameter are the same ones with the MIMLSVM algorithm and we selected the same values for them. Round parameter determines the number of boosting round which is 25 in our experiment.

\begin{figure}[t]
\center
\label{fig:figure2}
\includegraphics[width=3.7in]{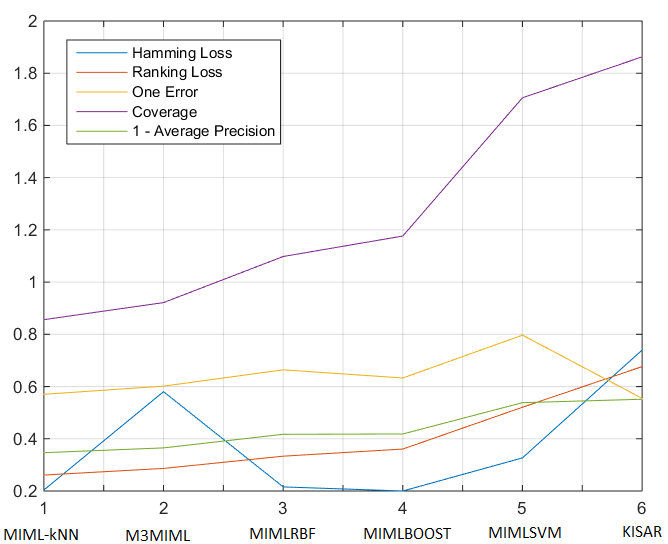}

\caption{\normalsize{Comparisons of the six algorithms on the dataset according to evaluation metrics.}}
\end{figure}

KISAR: This algorithm tries to find a relation between the bag of instances and class labels. The algorithm provides a mapping from a bag of instances to a feature vector, each bag of instances can be represented with these vectors so the vectors can be regarded as prototypes. According to the KISAR algorithm similarity of the prototypes of cases having relevant labels should be higher. Parameters are selected according to~\cite{li2012towards}.

M3MIML: This algorithm finds a linear model for each class. The outputs of each class is set according to maximum prediction of MIML example over the linear model. Then, the outputs on all possible classes are combined to define the margin of the MIML example. So, the relation between the instances and the labels of an MIML example are exploited by M3MIML. The cost parameter is chosen according to~\cite{zhang2008m3miml}.

\subsection{Evaluation Metrics}

In traditional supervised learning accuracy is often used as the performance evaluation criterion. However, while learning with multi-instance images associated with multiple labels, accuracy becomes less meaningful. Therefore, we used some other popular metrics that are more suitable for MIML such as hamming loss, one-error, coverage, ranking loss and average precision~\cite{zhou2012multi}.

\begin{itemize}
  \item Hamming loss: It evaluates how many times an example-label pair is misclassified, i.e., false alarm or misclassification. Since this is a loss function, optimal value for hamming loss is zero.
  \item One-error: It evaluates how many times the top-ranked label is not in the set of proper labels of the example. The performance is perfect when one-error = 0; the smaller the value of one-error, the better the performance.
   \item Ranking loss: It evaluates the average fraction of label pairs that are misordered for the example. The performance is better when the value of ranking loss is closer to zero.
  \item Coverage: It evaluates how many steps are need, on the average, to go down the list of labels in order to cover all the proper labels of the example. The performance is better when the value of coverage is smaller.
  \item Average precision: The average precision evaluates the average fraction of proper labels ranked above a particular label. The performance is perfect if the value of average precision equal to one.
\end{itemize}

We obtained predicted labels from six algorithms by using our dataset. We evaluate their effectiveness according to the metrics above. Results are shown in the table below:

\begin{tabular}{l*{10}{c}r}
Algorithms	             & h.l. & o.e. & r.l. & co. & a.p. \\
\hline
MIML-kNN 		& \textbf{0.203} & \textbf{0.261} & 0.570 & \textbf{0.856} & \textbf{0.653} \\
M3MIML   	  & 0.580 & 0.601 & \textbf{0.286} & 0.921 &  0.634 \\
MIMLRBF            & 0.210 & 0.339 & 0.640 & 1.130 &  0.586 \\
MIMLBOOST   	  & 0.200 & 0.632 & 0.360 & 1.176 &  0.581 \\
MIMLSVM           & 0.329 & 0.804 & 0.524 & 1.705 &  0.460 \\
KISAR   	  & 0.741 & 0.656 & 0.546 & 1.784 &  0.461 \\

\end{tabular}

\section{Discussion}
MIML-kNN obtained the best performance over other methods with respect to all evaluation criteria, although MIMLRBF and MIMLBOOST also achieved close performances according to only hamming loss. In addition, KISAR has the worst performance compared to others. Although the performances achieved in the experiments are not bad there is still room for improvement. For instance, we used shrank version of whole slide images by the rate of one over sixteen, when the original images are used the performance is expected to be improved. Since running time of such computation would be very high, we kept it like this.

We used LBP features due to their simplicity, robustness and efficiency. We did not consider any histopathological structure that a particular cancer type might have to keep it brief. A new research topic might be investigating such meaningful structures. Also, we extracted the LBP features only from nuclei channel. This might be improved by extracting features from H\&E staining, LAB channels etc. additionally. Moreover, label of other 200 pathologist that we did not use in this study can improve the performance as well as the other 120 unused histopathology images.

While applying algorithms to our dataset we chose parameters according to optimum values obtained in previous works. In contrast to the study~\cite{li2012towards}, which shows that KISAR and MIMLSVM achieved better performance than others, they are the worst two methods in our study in terms of average precision. This inconsistency might stem from the difference of datasets used.

Since each approach is robust to different problem characteristics, one can increase the performance by combining all these approaches in a smart manner. 

\section{Conclusion}
In conclusion, we compared six different multi-instance multi-label supervised learning algorithms on a subset of digiPATH dataset. Then, we used some popular MIML evaluation criteria to measure their performances. Considering that the dataset has limited number of samples and MIML learning is a highly complicated learning problem compared to traditional machine learning, we achieve significant performances almost in all approaches, especially MIML-kNN gave the best performance that is $\%65.3$ in average precision.

{\small
\bibliography{mybib}

\begin{thebibliography}{12}
\expandafter\ifx\csname natexlab\endcsname\relax\def\natexlab#1{#1}\fi
\providecommand{\url}[1]{\texttt{#1}}
\providecommand{\href}[2]{#2}
\providecommand{\path}[1]{#1}
\providecommand{\DOIprefix}{doi:}
\providecommand{\ArXivprefix}{arXiv:}
\providecommand{\URLprefix}{URL: }
\providecommand{\Pubmedprefix}{pmid:}
\providecommand{\doi}[1]{\href{http://dx.doi.org/#1}{\path{#1}}}
\providecommand{\Pubmed}[1]{\href{pmid:#1}{\path{#1}}}
\providecommand{\bibinfo}[2]{#2}
\ifx\xfnm\relax \def\xfnm[#1]{\unskip,\space#1}\fi
\bibitem[{Veta et~al.(2014)Veta, Pluim, van Diest, and
  Viergever}]{veta2014breast}
\bibinfo{author}{M.~Veta}, \bibinfo{author}{J.~Pluim}, \bibinfo{author}{P.~J.
  van Diest}, \bibinfo{author}{M.~A. Viergever},
\newblock \bibinfo{title}{Breast cancer histopathology image analysis: a
  review.},
\newblock \bibinfo{journal}{IEEE transactions on bio-medical engineering}
  \bibinfo{volume}{61} (\bibinfo{year}{2014}) \bibinfo{pages}{1400--1411}.
\bibitem[{Boyle et~al.(2008)Boyle, Levin et~al.}]{boyle2008world}
\bibinfo{author}{P.~Boyle}, \bibinfo{author}{B.~Levin}, et~al.,
  \bibinfo{title}{World cancer report 2008.}, \bibinfo{publisher}{IARC Press,
  International Agency for Research on Cancer}, \bibinfo{year}{2008}.
\bibitem[{Mercan et~al.(2014)Mercan, Aksoy, Shapiro, Weaver, Brunye, and
  Elmore}]{mercan2014localization}
\bibinfo{author}{E.~Mercan}, \bibinfo{author}{S.~Aksoy}, \bibinfo{author}{L.~G.
  Shapiro}, \bibinfo{author}{D.~L. Weaver}, \bibinfo{author}{T.~Brunye},
  \bibinfo{author}{J.~G. Elmore},
\newblock \bibinfo{title}{Localization of diagnostically relevant regions of
  interest in whole slide images},
\newblock in: \bibinfo{booktitle}{Pattern Recognition (ICPR), 2014 22nd
  International Conference on}, \bibinfo{organization}{IEEE},
  \bibinfo{year}{2014}, pp. \bibinfo{pages}{1179--1184}.
\bibitem[{Zhou and Zhang(2006)}]{zhou2006multi}
\bibinfo{author}{Z.-H. Zhou}, \bibinfo{author}{M.-L. Zhang},
\newblock \bibinfo{title}{Multi-instance multi-label learning with application
  to scene classification},
\newblock in: \bibinfo{booktitle}{Advances in Neural Information Processing
  Systems}, \bibinfo{year}{2006}, pp. \bibinfo{pages}{1609--1616}.
\bibitem[{Yang(2005)}]{yang2005review}
\bibinfo{author}{J.~Yang}, \bibinfo{title}{Review of multi-instance learning
  and its applications}, \bibinfo{type}{Technical Report}, Tech. Rep,
  \bibinfo{year}{2005}.
\bibitem[{Zhang and Zhou(2008)}]{zhang2008m3miml}
\bibinfo{author}{M.-L. Zhang}, \bibinfo{author}{Z.-H. Zhou},
\newblock \bibinfo{title}{M3miml: A maximum margin method for multi-instance
  multi-label learning},
\newblock in: \bibinfo{booktitle}{Data Mining, 2008. ICDM'08. Eighth IEEE
  International Conference on}, \bibinfo{organization}{IEEE},
  \bibinfo{year}{2008}, pp. \bibinfo{pages}{688--697}.
\bibitem[{Bishop et~al.(1995)}]{bishop1995neural}
\bibinfo{author}{C.~M. Bishop}, et~al.,
\newblock \bibinfo{title}{Neural networks for pattern recognition}
  (\bibinfo{year}{1995}).
\bibitem[{Zhang and Wang(2009)}]{zhang2009mimlrbf}
\bibinfo{author}{M.-L. Zhang}, \bibinfo{author}{Z.-J. Wang},
\newblock \bibinfo{title}{Mimlrbf: Rbf neural networks for multi-instance
  multi-label learning},
\newblock \bibinfo{journal}{Neurocomputing} \bibinfo{volume}{72}
  (\bibinfo{year}{2009}) \bibinfo{pages}{3951--3956}.
\bibitem[{Zhang(2010)}]{zhang2010k}
\bibinfo{author}{M.-L. Zhang},
\newblock \bibinfo{title}{A k-nearest neighbor based multi-instance multi-label
  learning algorithm},
\newblock in: \bibinfo{booktitle}{Tools with Artificial Intelligence (ICTAI),
  2010 22nd IEEE International Conference on}, volume~\bibinfo{volume}{2},
  \bibinfo{organization}{IEEE}, \bibinfo{year}{2010}, pp.
  \bibinfo{pages}{207--212}.
\bibitem[{Li et~al.(2012)Li, Hu, Jiang, and Zhou}]{li2012towards}
\bibinfo{author}{Y.-F. Li}, \bibinfo{author}{J.-H. Hu},
  \bibinfo{author}{Y.~Jiang}, \bibinfo{author}{Z.-H. Zhou},
\newblock \bibinfo{title}{Towards discovering what patterns trigger what
  labels},
\newblock in: \bibinfo{booktitle}{Twenty-Sixth AAAI Conference on Artificial
  Intelligence}, \bibinfo{year}{2012}.
\bibitem[{Ruifrok and Johnston(2001)}]{ruifrok2001quantification}
\bibinfo{author}{A.~C. Ruifrok}, \bibinfo{author}{D.~A. Johnston},
\newblock \bibinfo{title}{Quantification of histochemical staining by color
  deconvolution.},
\newblock \bibinfo{journal}{Analytical and quantitative cytology and
  histology/the International Academy of Cytology [and] American Society of
  Cytology} \bibinfo{volume}{23} (\bibinfo{year}{2001})
  \bibinfo{pages}{291--299}.
\bibitem[{Zhou et~al.(2012)Zhou, Zhang, Huang, and Li}]{zhou2012multi}
\bibinfo{author}{Z.-H. Zhou}, \bibinfo{author}{M.-L. Zhang},
  \bibinfo{author}{S.-J. Huang}, \bibinfo{author}{Y.-F. Li},
\newblock \bibinfo{title}{Multi-instance multi-label learning},
\newblock \bibinfo{journal}{Artificial Intelligence} \bibinfo{volume}{176}
  (\bibinfo{year}{2012}) \bibinfo{pages}{2291--2320}.

\end{thebibliography}
\bibliographystyle{model1-num-names}
}

\end{document}